\documentclass[submission,copyright,creativecommons]{eptcs}
\providecommand{\event}{ICLP/LPNMR-DC 2024} 

\usepackage{iftex}
\usepackage{multirow}
\usepackage[frozencache,cachedir=.]{minted}

\ifpdf
  \usepackage{underscore}         
  \usepackage[T1]{fontenc}        
\else
  \usepackage{breakurl}           
\fi

\def\titlerunning{Commonsense Reasoning-Aided Autonomous Vehicle Systems}
\def\authorrunning{K. Kimbrell}

\title{Commonsense Reasoning-Aided Autonomous Vehicle Systems}
\author{Keegan Kimbrell
\institute{The University of Texas at Dallas\\ Richardson, USA}
\email{keegan.kimbrell@utdallas.com}
}
\begin{document}
\maketitle

\begin{abstract}
Autonomous Vehicle (AV) systems have been developed with a strong reliance on machine learning techniques. While machine learning approaches, such as deep learning, are extremely effective at tasks that involve observation and classification, they struggle when it comes to performing higher level reasoning about situations on the road. This research involves incorporating commonsense reasoning models that use image data to improve AV systems. This will allow AV systems to perform more accurate reasoning while also making them more adjustable, explainable, and ethical. This paper will discuss the findings so far and motivate its direction going forward.
\end{abstract}

\section{Introduction}
For both academic and industry research, AV technology has seen incredible advances since the introduction of computer vision-focused systems in the 1980's \cite{bimbraw2015autonomous}. Here, this paper will provide some formal definitions for autonomous vehicles that it will use throughout this writing. SAE International defines autonomous vehicles into six different levels based on the level of automation, with level 0 being no automation and level 5 being full driving automation \cite{on2021taxonomy}. Despite AV research being a well-explored field, there are still no level 5, or fully autonomous, vehicles. This is largely due to imperfections in computer vision systems and the complexity of more complicated driving tasks that require a human driver to be present. For a safety-critical system, such as AV systems, minor mistakes cannot be afforded. To this end, it is important that the AV system can make safe and rational decisions based on accurate interpretations about its surroundings.

There are several technologies that are used in the perception side of AV systems, such as Light Detection and Ranging (LiDAR) systems and camera-based systems. These systems are coupled with deep learning techniques such as Convolutional Neural Networks (CNNs), which are used to classify sensor data \cite{quito2023compare}. However, like all machine learning systems, it is always possible for misclassifications to occur due to noise, scenarios outside of the training data, degradation of sensing equipment, and other external factors. Because of this, AV systems should move towards using a hybrid AI system, or AI that combines deep learning with logical reasoning, to help mitigate the failures and shortcomings of solely deep learning-based approaches.

There are two types of systematic thinking proposed by Kahneman in 2011 \cite{kahneman2011thinking}. The first is "System 1", which is fast, instinctive, and emotional thinking. The second is "System 2", which is slow, deliberative, and logical. For a human driver, we use both systems when we are in a driving scenario. Identifying objects around us and minor driving actions are done quickly using System 1 thinking. However, when we encounter an unfamiliar or dangerous scenario, we use System 2 thinking to determine a safe way to navigate the situation. In an optimal hybrid AV system, fast System 1 tasks such as perception and classification should be handled by deep learning, and slow System 2 tasks should be handled by commonsense reasoning. The reasoning system can also be used to perform a more deliberative analysis of the sensor data. This is similar to a human driver who realizes that they misinterpreted an object on the road and looks closer to figure out what it is. 

Commonsense reasoning is a method for modeling the human way of thinking about the world around us using default rules and exceptions \cite{gelfond2014knowledge}. In the context of driving scenarios, we can understand this as our default understanding of traffic laws and scenarios. For example, if we drive up to a crosswalk, by default, we know that we have to stop when there are pedestrians waiting to cross. However, there may be an exceptional scenario that breaks this rule, such as if we discover that the pedestrians do not actually intend to cross. In this case, a human driver will still stop and slowly think about the situation and confirm that the pedestrians do not want to cross before making the potentially unsafe decision of driving forward. This research is focused on modeling commonsense reasoning and combining it with current AV techniques to create safer and more reasonable autonomous vehicles. 

This experiment proposes a framework for improving AV systems by attaching commonsense layers that use image data to provide feedback to the deep learning layers for various tasks. With this approach, we can write commonsense reasoning models that can perform optimizations, safety checks, and explanations for autonomous vehicles. By keeping the commonsense reasoning model in a separate layer, we can even use this approach to improve existing AV systems. Furthermore, this approach is not encumbered by a mandatory and expensive training process. The commonsense reasoning model can be modeled and updated with rules generated from domain experts, allowing us to easily stay up-to-date on new laws, ethical standards, and regulations. Currently, the commonsense model employs collective behaviors, or the actions of nearby vehicles, to determine the state of the road around us.

\section{Related Work}
There have been many other works that incorporate symbolic reasoning into deep learning, computer vision, and autonomous vehicle models. Suchan et al. explore commonsense reasoning-based approaches such as an integrated neurosymbolic online abduction vision and semantics-based approach for autonomous driving \cite{suchan2021commonsense, suchan2020driven}. These techniques are primarily focused on integrating with the perception model using answer set programming (ASP), a nonmonotonic reasoning system using stable models \cite{gelfond2014knowledge,lifschitz2019answer}. While their framework is similar, this approach is more decoupled from the vision models, allowing us to show improvements on existing AV models.

Neurosymbolic AI, AIs that integrate symbolic and neural network-based approaches \cite{hitzler2022neuro}s, have been applied towards autonomous driving as well. For safety-critical systems, such as autonomous driving, neurosymbolic techniques can improve compliance with guidelines and safety constraints \cite{sheth2023neurosymbolic}. Anderson et al. propose a neurosymbolic framework that incorporates symbolic policies with a deep reinforcement learning model \cite{anderson2020neurosymbolic}. They assert that this approach can improve the safety of reinforcement learning approaches in safety-critical domains, including autonomous vehicles. These systems are related to this research in the sense that both are using symbolic methods to improve existing deep learning-based systems. However, this research is different in that it is using commonsense reasoning as the proposed symbolic model and that, while it is being used to improve on a deep learning model, it is a different layer that is generated separately. While autonomous vehicles and computer vision technologies are primarily deep learning-based, this approach could be used to improve upon reinforcement learning-based, other non-neural machine learning-based, or even search-based vehicles.

A framework created earlier, AUTO-DISCERN \cite{kothawade2021auto}, proposes a goal-directed commonsense reasoning ASP system that makes driving decisions based on the observations of the environment. This research is an extension of this approach by creating a commonsense reasoning model that makes safe decisions and reasons over a road scenario. This experiment pushes it farther by incorporating the model with an AV system and using the commonsense model to improve aspects of autonomous driving.

\section{Research Goals}
The major goal of creating an AV system of higher level autonomy using commonsense reasoning can be broken down into various tasks relating to where in the AV system we inject the reasoning:
\begin{itemize}
\item \textbf{Perception and Classification}: Use commonsense reasoning and knowledge to model the sensor data and optimize the classifications. We can also inject commonsense reasoning into the training process itself to create a more connected and explainable model.
\item \textbf{Safe Decision Making}: We can model rules for the AV system so that it will always make intelligent and safe decisions that still move it towards its desired goal. This is important for making an ethical system that complies with traffic laws. 
\item \textbf{Complicated Tasks}: Complicated tasks may be outside of the training data for an AV system, such as navigating a road after the results of a hurricane. We can create reasoning models that can handle multiple unknown scenarios safely. Furthermore, it is easier to model these niche scenarios using reasoning since there is often a strong bias against such scenarios in the existing training data for AV systems, and they are difficult to capture using just deep learning.
\end{itemize}

Each of these tasks separately will improve the effectiveness of future AV systems, and if success is found in each task, then they can be combined to create an autonomous vehicle with a higher level of autonomy than existing systems.

\section{Preliminary Results}
The current focus of these recent experiments has been using commonsense reasoning and knowledge to optimize the classifications of the computer vision model through consistency checking (first and second tasks). The system uses a Prolog \cite{clocksin2003programming} commonsense reasoning model to check if the classifications being made by the computer vision model are consistent with each other, particularly if the behavior of nearby vehicles is consistent with the current road scenario. For example, if a traffic light at an intersection is red, then vehicles in that lane should be stopped. We define the group actions of nearby vehicles as \textit{collective behaviors}. If the rules about the collective behaviors are not consistent with the observed objects, then the system adjusts the classifications of objects around the AV system to fix the scenario. This system emulates the human process of reasoning about a road situation by observing surrounding vehicles. In this approach, we test over misclassified traffic light colors and unobserved road obstacles.

To accomplish this, the system takes objects from the computer vision model's output and converts them into facts. For example, the following facts represent the information about nearby vehicles and intersections:

\smallskip
\begin{minted}{prolog}
property(vehicle, Frame, Object_id, 
    Action, VelocityX, VelocityY, Rotation, 
    Coordinate1X, Coordinate1Y, 
    Coordinate2X, Coordinate2Y).
property(intersection, Frame, Object_id, 
    Coordinate1X, Coordinate1Y, 
    Coordinate2X, Coordinate2Y).
    vehicles(Frame, Vehicles).
\end{minted}
\smallskip

These facts are treated as knowledge about our current scenario. The system also contains another Prolog program that performs commonsense reasoning over road scenarios. These rules define how nearby vehicles, or collective behaviors, should act around traffic lights and obstacles.

\smallskip
\begin{minted}{prolog}
false_negative_light(Frame):- 
    property(intersection, Frame, _, _, _, _, _), 
    collective_{up/down/left/right}(Frame).
\end{minted}
\smallskip

This rule is a basic example of how the system would detect misclassifications about the presence of red traffic lights. The rule will evaluate to true, meaning that the AV system fails to detect a red traffic light (negative) when there actually is one (positive), when the AV system detects a collective behavior moving across an intersection in front of it. This is similar to how a human driver can figure out a traffic light is red even if it appears to be green based on the behavior of nearby vehicles. In addition to rules concerning traffic lights, the commonsense reasoning program also contains rules about how vehicles behave when near obstacles.

The following are some of the results from the experiments performed so far, which show the results of this system when identifying the color of an incoming traffic light and road obstacles using the collective behaviors of nearby vehicles. Both experiments were performed over recorded datasets from the Car Learning to Act simulator (CARLA) \cite{dosovitskiy2017carla}. 
\begin{table}[]
\centering
\def\arraystretch{1.5}
\begin{tabular}{|c|c|c|c|c|}
\hline
\multirow{2}{*}{CARLA Traffic Lights} & \multicolumn{4}{c|}{Metrics}                                                                           \\ \cline{2-5} 
                       & \multicolumn{1}{l|}{Accuracy} & \multicolumn{1}{l|}{Precision} & \multicolumn{1}{l|}{Recall} & F-Score \\ \hline
Town 1 100 NPCs Logic           & .9632        & .9663     & .9942      & .98 \\ 
Town 1 100 NPCs Baseline        & .479         & .6579     & .145       & .237 \\   
Town 1 100 NPCs Combined        & .9547        & .9297      & .9942     & .9609  \\  \hline
Town 1 200 NPCs Logic           & 1            & 1         & 1          & 1     \\ 
Town 1 200 NPCs Baseline        & .7634        & .5        & .4091      & .45   \\   
Town 1 200 NPCs Combined        & .8387        & .64       & .7272      & .6809 \\  \hline
\end{tabular}
\caption{Results of commonsense reasoning, baseline deep learning, and combined hybrid models for traffic lights. The logic model is only evaluated over frames in which there are collective behaviors, and the baseline and combined models are evaluated over all frames.}
\label{table:1}
\end{table}

Table \ref{table:1} shows the accuracy of the logic model and baseline computer vision model when it comes to identifying the color of traffic lights at intersections. The dataset is generated from two different recordings, about a couple minutes long each. Each recording was done using the same CARLA map (Town 1) with two different vehicle population densities (100 and 200). The image data contained inclement weather conditions that were outside of the training data. Due to this, the baseline model struggles to maintain high accuracy when identifying the color of the traffic light. 

The accuracy of the commonsense reasoning model is evaluated over eligible frames in the data, meaning images within the data that fulfill the default rules in our model. For these eligible frames, the commonsense reasoning model maintains high accuracy. When the reasoning model is combined with the baseline model and evaluated over the whole dataset, we can see a significant increase in all metrics over the baseline model. The increase in performance varies, as it depends strongly on how many frames are eligible for the commonsense reasoning to perform a correction. This is why there is a greater increase in accuracy for the first dataset as opposed to the second. Despite this, the experiment so far has demonstrated that this approach is an effective optimizer for this scenario. 

\begin{table}[]
\centering
\def\arraystretch{1.5}
\begin{tabular}{|c|c|c|c|c|}
\hline
\multirow{2}{*}{CARLA Obstacles} & \multicolumn{4}{c|}{Metrics}                                                                           \\ \cline{2-5} 
                       & \multicolumn{1}{l|}{Accuracy} & \multicolumn{1}{l|}{Precision} & \multicolumn{1}{l|}{Recall} & F-Score \\ \hline
Town 3.0 Logic              & 1   & 1   & 1      & 1 \\ 
Town 3.0 Baseline           & .4943   & 1   & .4943   & .6615 \\ 
Town 3.0 Combined           & 1   & 1   & 1      & 1 \\ \hline
Town 3.1 Logic              & 1   & 1   & 1      & 1 \\ 
Town 3.1 Baseline           & .93   & 1   & .93      & .9636 \\ 
Town 3.1 Combined           & 1   & 1   & 1      & 1 \\ \hline
\end{tabular}
\caption{Results of the commonsense reasoning model for obstructions for the logic, baseline, and hybrid combined models.}
\label{table:2}
\vspace{-0.2in}
\end{table}

The results from Table \ref{table:2} show a similar result for a different scenario. In this dataset, scenarios are much shorter (around 30 seconds) and are used to evaluate predictions about incoming obstacles that are blocking a lane of traffic. This is a task that deep learning models can struggle with since they rely entirely on image or sensor data. If a large vehicle is completely obstructing the view of the obstacle, the deep learning system will struggle heavily to identify it. This, however, is not an issue for the commonsense reasoning system. The results show that as long as there are vehicles nearby for us to observe, we can always determine an obstacle blocking a lane. The accuracy of the deep learning model depends heavily on how well it can see the obstruction, which is what leads to the results seen in the table. 

\section{Conclusion and Future Work}
While a lot of progress has been made in research for AV technology, we are still far away from achieving a fully autonomous vehicle. This is because of an overreliance on deep learning techniques. Proposed here is a pipeline towards a fully autonomous vehicle by incorporating commonsense reasoning into various aspects of the AV system. The results so far demonstrate the effectiveness of this approach.

This work will be extended by exploring new techniques to improve the applicability and efficiency of this approach. This approach can be improved with evaluations from real-world datasets, such as KITTI or NuScenes \cite{geiger2013vision,caesar2020nuscenes}, the use of more powerful logic technologies, such as answer set programming, and the exploration of efficient ways to construct and invoke commonsense reasoning models. It will also benefit from the consideration of new ways to combine commonsense reasoning into AV systems, such as employing more neurosymbolic-based methods like injecting commonsense into the training of the deep learning model.

Going forward, the techniques shown for autonomous vehicles can be applied to other domains. These approaches focus on using commonsense reasoning models that use the images from the autonomous vehicle to improve the system. This can be viewed as a form of visual question answering (VQA \cite{antol2015vqa}) and can be applied to other domains. Future work will be about the knowledge extraction and reasoning from images used in this experiment and demonstrate its effectiveness in various applications, including autonomous vehicles.

\nocite{*}
\bibliographystyle{eptcs}
\bibliography{ICLPDC}
\end{document}